# Best Practices for Text Annotation with Large Language Models

Petter Törnberg

Institute for Language, Logic and Computation (ILLC), University of Amsterdam.

p.tornberg@uva.nl.

**Abstract**

Large Language Models (LLMs) have ushered in a new era of text annotation, as their ease-of-use, high accuracy, and relatively low costs have meant that their use has exploded in recent months. However, the rapid growth of the field has meant that LLM-based annotation has become something of an academic Wild West: the lack of established practices and standards has led to concerns about the quality and validity of research. Researchers have warned that the ostensible simplicity of LLMs can be misleading, as they are prone to bias, misunderstandings, and unreliable results. Recognizing the transformative potential of LLMs, this paper proposes a comprehensive set of standards and best practices for their reliable, reproducible, and ethical use. These guidelines span critical areas such as model selection, prompt engineering, structured prompting, prompt stability analysis, rigorous model validation, and the consideration of ethical and legal implications. The paper emphasizes the need for a structured, directed, and formalized approach to using LLMs, aiming to ensure the integrity and robustness of text annotation practices, and advocates for a nuanced and critical engagement with LLMs in social scientific research.

**Significance statement:** Instruction-tuned Large Language Models have seen rapid and widespread adoption in academic research in recent months. As LLMs redefine the landscape of textual analysis with their efficiency and accessibility, the absence of shared standards for their application has raised significant concerns regarding the quality, bias, and reliability of resulting work. By proposing a comprehensive set of guidelines, this paper provides a framework to enable rigorous, reproducible and ethical LLM-based text annotation, ensuring that this powerful tool serves as a catalyst for robust and responsible social scientific inquiry.

**Introduction**

The recent year has seen instruction-tuned Large Language Models (LLM) emerge as a powerful new tool for text annotation. These models are capable of text annotation based on instructions written in natural language – so called *prompts* – thus obviating the need of training of models on large sets of manually classified training data (Wei, Bosma, et al., 2022). The ease-of-use, high accuracy, and relatively low costs of LLMs have meant that their use has exploded in recent months – appearing to represent a paradigm shift in text-as-data by enabling even researchers with limited knowledge in computational methods to engage in sophisticated large-scale textual analysis (Gilardi et al., 2023; Rathje et al., 2023; Törnberg, 2023). Such use is helped by the quickly growing availability of simple how-to guides with code examples on how LLMs can be employed for text annotation (Törnberg, 2024).

While LLMs are enabling exciting new research and empowering researchers to carry out sophisticated studies, the rapid growth of the field is not without problems. LLM-based text annotation has become something of an academic Wild West, as the lack of established standards has meant that both researchers and reviewers lack benchmarks for LLM-based annotation, leading to risks of low-quality research and the publishing of invalid results. LLMs fit poorly into our existing frameworks for thinking about research methods: many of the lessons from machine learning are obsolete, and while LLMs can be eerily similar to working with a human coder, this model can be similarly misleading. While seemingly simple to use, the models are black boxes, and prone to bias, misunderstandings, and unreliable results – leading some researchers to warn against at all using the models for annotation (Kristensen-McLachlan et al., 2023; Ollion et al., 2023). The models raise important questions about bias, calibration, and validation, and the field is thus in need of common standards for what constitutes acceptable and recommended research practices.

This brief paper seeks to contribute to addressing this need by suggesting a set of standards and best practices for how LLMs can be employed reliably, reproducibly, and ethically. The paper argues that, while LLMs can indeed be prone to display bias and unreliable results, we should not reject their use altogether – instead, we should manage their potential weaknesses by bringing them into a rigorous annotation process. The paper draws on existing research published using LLMs, the authors own extensive work in the field, and discussions with scholars working in the field. The paper targets both researchers seeking advice on how to use LLMs in a rigorous and reliable way, and reviewers seeking standards for evaluating research.

We will cover the following eight points: (1) choose an appropriate model, (2) follow a systematic coding procedure, (3) develop a prompt codebook (4) validate your model, (5) engineer your prompts, (6) specify your LLM parameters, (7) discuss ethical and legal implications, (8) examine model stochasticity, (9) consider that your data may be in the training data.

**1. Choose an appropriate model**

The choice of which LLM to use is one of the most central decisions in LLM-based text analysis (Yu et al., 2023). There are now a large and diverse set of models to choose from, ranging from small open-source local models that can be run on a phone to large platformed models accessible through a web interface or API – so called AIAAS (Artificial Intelligence As A Service).

Yet, most existing studies using LLMs for text annotation have employed one of OpenAI's proprietary models – either GPT3.5 or GPT4, and few have offered any explicit motivation for their choice. The popularity of the OpenAI models is likely due to their sophistication, relatively low price, ease-of-use, and that they are the most well-known models – but they also come with several important problems. First, the ChatGPT models have been shown to change over time without notice, giving different results to the same instructions as a result to changes in the backend (Chen et al., 2023). While the API provides access to stable models, these are deprecated after a relatively short time, making reproducibility an impossibility. Second, as it is not known what data these models are trained on, the OpenAI models do not pass even a low bar of transparency (Liesenfeld et al., 2023). Third, using a model through an API can be problematic in terms of ethics and legal consideration for certain data, and the current advice is that OpenAI models should not be used with any proprietary, secret, or confidential data (Ollion et al., 2024; Spirling, 2023).

The choice of model should be explicitly argued for, and we can point to six general factors that should be taken into account when selecting which LLM for annotation:

1. **Reproducibility**: The results can be replicated by others using the same data and methodology, ensuring the results are consistent and reliable. To ensure reproducibility, use a fixed version of the LLM throughout the project, document the version, and ensure that the model will be available for future use.
2. **Ethics and legality:** The model should respect ethical and legal standards, including considerations of privacy, not storing research data, and compliance with relevant data privacy regulations.
3. **Transparency**: The methodologies, data sources, assumptions, and limitations of the model should be clearly documented and accessible for scrutiny.
4. **Culture and Language**: The LLM should adequately support the language(s) and cultures of your textual data. Some models are more proficient in certain languages than others, which can influence the quality of the annotations – and even bias your findings if your corpus includes several languages. Specifically, many models are English-centric, which can result in lower performance on other languages and cultures (Ollion et al., 2024).
5. **Scalability**: Ensure that the model can handle the size of your relevant data material in terms of costs and time. The speed of offline models depends largely on the available hardware, whereas it for API-based models depends on their rate limits and costs. (If you need to classify large amounts of data, it may be worth considering using a semi-supervised model trained on data annotated by the LLM. While this adds an additional step, such models tend to be faster and are possible to run on an average laptop, thus allowing processing large quantities of data.)
6. **Complexity**: Ensure that the model has the capacity to handle the complexity of the task, for instance relating to advanced reasoning or parsing subtle latent meaning. Challenging analysis tasks and long prompt instructions may require larger and more sophisticated models, such as GPT4.0, that are capable of higher levels of reasoning and performance on benchmark tasks.

The best practice is to use an open-source model for which the training data is publicly known. It should be noted that not all downloadable models can be considered open-source models, as models vary significantly in terms of their openness of code, training data, model weights, licensing, and documentation – and it is therefore important to compare the models based on existing benchmarks for openness (Liesenfeld et al., 2023). The models also vary significantly in their capacity for text annotation. Some open source models have been found to yield results comparable to those of ChatGPT for certain task (Alizadeh et al., 2023; Weber and Reichardt, 2023). To compare and select an appropriate model, there are several benchmarks and leaderboards that provide an overview of the capacities of the quickly changing landscape of available models (Bommasani et al., 2023; Chia et al., 2023; HuggingFace, 2024). The choice of model should also take into account that guard-rail tuning – in which the models are trained to avoid controversial statements (Fernandes et al., 2023; Ziegler et al., 2019) – can be problematic for certain annotation tasks, as the models may refuse to annotate particular controversial issues, for instance that COVID19 vaccine resistance is more associated to one United States political party than the other (Törnberg, 2023).

Custom fine-tuning – that is, training the model using labelled data on your specific annotation task – can further improve the results of text annotation, and can enable you to achieve excellent results with an open-source model, even with as few as 50-100 manually labeled examples.

If possible, the model should be hosted on your own infrastructure instead of relying on cloud-based APIs. Hosting the model yourself gives you complete control over the model version and updates, as well as over how the model handles any sensitive or confidential information, and makes your work replicable. While self-hosting is not available for all models, it can be surprisingly easy, cheap, and faster than API-based models, depending on your available hardware and the annotation task at hand. Ideal practice also involves assessing whether your results can be reproduced using several models, thereby showing that the prompt and results are robust to details of implementation. You should in general *never* use LLMs for annotation through their web interface, as these interfaces do not allow setting parameters, version control, and do not provide sufficient privacy or copyright provision – the data you provide is often kept and used for training models.

However, the best model ultimately depends on the task at hand, and it should be acknowledged that there are often trade-offs. It may, for instance, not be possible to use a smaller open-source model for complex tasks, and the researcher may thus be forced to use a model such as GPT-4. In choosing the model, it is useful to look at what instructions the model was tuned on, and how the model scores on benchmarks that are relevant for your domain of application (Chang et al., 2024). While it likely that we will soon see the development and standardization of academic-led open source academic LLMs specifically developed for data annotation, which will help resolve these tradeoffs (Spirling 2023), the bottom-line is thus that *the choice of model must always be motivated and argued for on the basis of explicit quality standards*.

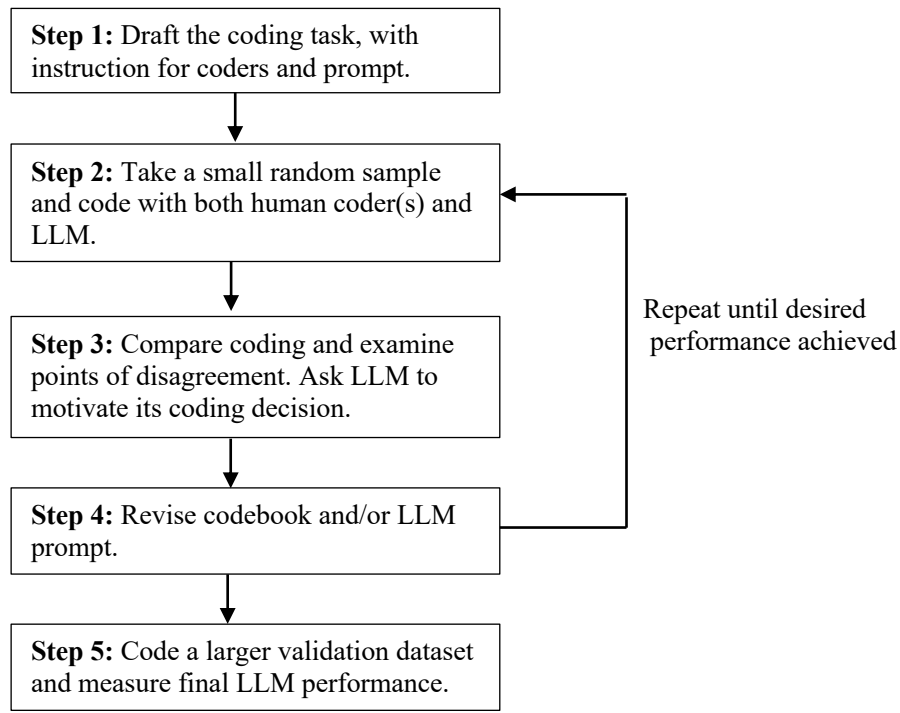

Figure 1: Example of a systematic coding procedure.

## 2. Follow a Systematic Coding Procedure

Text annotation is rarely merely a straight-forward technical task but involves the challenging work of defining the meaning of concepts. There are almost always difficult boundary cases, which tend to become obvious only when engaging with the data, and some level of subjectivity is hence inevitable in coding. It is central to strive for consistency and transparency in how this subjectivity is managed. Since LLMs can be fallible, unreliable, biased, and can misunderstand instructions and produce unexpected results (Ollion et al., 2023), it is important that the LLM is integrated into a systematic coding procedure that handles these issues and aligns their coding with the intended task. Such procedures are already well-established when it comes to groups of human coders, and the LLM can successfully be brought into such a process.

Annotation work is generally organized as an iterative coding process (Chun Tie et al., 2019; Glaser and Strauss, 2009): coders start with a set of texts, discuss discrepancies, refine the guidelines, and then proceed with the next set of texts. Such calibration sessions, where coders align their understanding and application of the guidelines, are crucial for maintaining consistency. When coding with an LLM, the development of the prompt is simply brought into this loop – simultaneously developing coding instructions and the LLM prompt. Once the LLM reached sufficient agreement with the human coders, it can be used to code the full material.

Taking this approach, you can calculate the reliability both across the human coders and with the LLM. This allows assessing how well the LLM performs the task compared to human coders, and track the convergence between the coders and the LLM. Ideally, the LLM should approach the reliability achieved among the human coders.

1. **Define the concept:** It is important to come in with an explicitly articulated idea of the concept you are trying to capture, to avoid being overly influenced by the interpretations of the LLM. Write up a first description of the task at hand in the codebook, with instructions for both the human coder(s) and for the LLM. Make the prompt clear, unambiguous, and specific, using direct and instructional language. (While the human instructions and the LLM prompt should generally be similar, it is usually beneficial to provide separate instructions.)
2. **Code a small training dataset:** Have the human coders code a small representative dataset to enable testing your prompts, and use the LLM to annotate the same data.
3. **Examine points of disagreement:** Check the agreement between coders, and between the coders and the LLM. Discuss cases where coders disagree amongst each other, and on cases where the LLM disagreed with the coders. Ask the model to motivate its annotations for these cases and compare with the motivations of the human coder – as this can be a useful tool for sharpening your operationalization. (To do so, feed the model the question again, plus its response, followed by a request to motivate its answer. Note that the usefulness of these responses depends on the model; some open-source models provide only low-quality motivations.) At the same time, it is important to be self-critical and reflexive, and not be overly swayed by the model's interpretations.
4. **Refine the codebook and prompt:** Make necessary adjustments to the instructions of either the human coders, of the prompt, or both. The human coders should not be considered ground truth: you may find that the LLM's interpretation was superior to the human coder. When used mindfully, the LLM can be a powerful tool for conceptual work.

5. **Repeat:** Return to step 2. Continue this process until the desired output quality is achieved.
6. **Validate:** Code the validation dataset, and measure the final performance of the LLM (see *4. Validate your model*).

Note that the process described above is merely an example and may need to be adapted to the specific needs of the project. If the zero-shot prompt is not giving adequate results, it can be useful to add few-shot examples. If the results are still inadequate, consider fine-tuning the model based on labeled training data.

## 3. Develop a Prompt Codebook

Best practice involves developing a *prompt codebook* with annotation guidelines for the human coders combined with detailed description of the prompts and LLM settings. The coding guidelines should, as always (Glaser and Strauss, 2009), be detailed instructions with clear definitions of the coding categories, examples of text corresponding to each category, and instructions on how to handle ambiguous cases. A coder (human or LLM) that reads the codebook should have sufficient information to code a given text, with minimal disagreement between coders. The codebook should simultaneously describe the corresponding prompts and parameters for the LLM, providing all details necessary to reproduce the LLM coding. This enables full reproducibility of both the manually coded validation data and the LLM coding. Note that the prompt should be considered tailored to the model used for its development: applying the same prompt to a different LLM may produce different results, even with models of similar parameter size (Sanh et al., 2022). If you finetune your model for your specific annotation task, the data used should be provided.

## 4. Validate your model

While studies have shown that LLMs can in certain cases outperform expert human coders in text annotation (Törnberg, 2023), LLM performance has been found to be highly contingent on both the dataset and the type of annotation task (Kristensen-McLachlan et al., 2023). There is also an always-present risk for biases in the models. Hence, it is always necessary to carefully validate the models on a task-by-task basis (Pangakis et al., 2023). Validation is, in short, a basic requirement for all publications using LLMs.

Validation usually consists of manually labelling a sufficient number of texts and ensuring that the labels correspond to a sufficient degree with the model results (Karjus, 2023). When the LLM is used to provide data for a supervised model, the validation data can be used both to validate the results of the LLM, and of the supervised model.

There are several requirements for satisfactory validation:

- **The validation must take place *after* the annotation prompt has been finalized**: it is not acceptable to use the validation data to improve the prompts, as this may lead to falsely reporting higher precision.
- **The validation dataset needs to be sufficiently large:** The exact amount of validation data needed depends on several factors, such as the number of categories and the balance of categories. If the categories are imbalanced (some categories have many more examples than others), you might need more data to ensure that the model performs well on the less-represented categories. The practical minimum is to have at least 20-30 samples of each category for a basic level of confidence in the performance metrics, but more is generally better. For high-stakes applications, you may need significantly more to ensure robustness. (For a precise determination, consider performing a power analysis.)
- **Use appropriate performance metrics:** Accuracy – i.e., correct answers divided by total answers – *is generally not a sufficient measure* to evaluate model performance, as it can be highly misleading, in particular for imbalanced datasets. (If, for instance, one of your categories represents 90% of the population, then a model which puts everything that category will achieve a seemingly impressive accuracy of 90%.) What measure is appropriate however depends on the task at hand. For classification, measures such as *F1 Score* (usually together with *precision* and *recall*), *weighted-F1 score*, *ROC-AUC*, or *Cohen's Kappa* can be appropriate, whereas correlation-based measures, *MAE* or *MSE* can be more relevant when the model is annotating numeric values. In short, you need to argue for why your measure is the most appropriate choice, and it is in in practice often beneficial to use a combination of these metrics to get a comprehensive understanding of different aspects of the model's performance.
- **Consider comparing with human performance:** Certain tasks are inherently more challenging than others, and the acceptable performance level will vary depending on the task at hand. Calculating the performance of human coders though e.g., an inter-coder reliability score, can provide a useful benchmark for evaluating of the performance of a model.

- **Consider any subsets of the data:** If your dataset includes several subsets for which the model's capacity may vary, for instance different languages or cultural contexts, they need to be separately validated as the model may vary in its precision for each group.
- **Examine and explain failures:** LLMs are not human, and their performance may thus vary in unexpected ways – possibly involving a problematic misinterpretation of the concept. It may be that the model fails for specific interpretative tasks, that may lead to biases in the downstream analysis, without being visible in the performance metrics. You should therefore always examine the failures and provide a description of the weaknesses of the model, and why they do not undermine the validity of downstream results.

While it is likely that we will soon see certain prompts and models become well-established for certain analysis tasks, the general advice is that any automated annotation process using LLMs *must* validate their LLM for their specific prompt, settings, and data. Rigorous validation is the most important step in using LLMs for text annotation.

**5. Engineer your prompts**

One of the main implications of the use of LLMs for text annotation is the emergence of the task of *prompt engineering*: developing instructions that guide the LLM. While prompts are written in natural language and do not require technical skills per se, there can be huge differences in performance depending on details of how the prompt is written. Prompt engineering is hence becoming an important social scientific skill (White et al., 2023). Writing effective prompts can require significant effort, with multiple iterations of modification and testing (Jiang et al., 2020). While many prompting techniques have been developed, there is still limited theoretical understanding of why a particular technique is suited to a particular task (Zhao et al., 2021).

While previous advances in computational methods within the social sciences has tended to require sophisticated technical skills, prompt engineering requires other social scientific skills, such as theoretical knowledge, communication ability, and capacity for critical thinking. The process of developing prompts can furthermore be a useful way of deepening our understanding of social scientific concepts. Prompt engineering can in this sense therefore be thought of as a new type of – or even extension of – qualitative social science (Karjus, 2023). This paper will not provide a complete introduction to prompt engineering, as such guides are already readily available (e.g., OpenAI, 2024; Saravia, 2022), but will provide some important general advice.

- **Structured prompts:** An annotation prompt should contain the following elements: *context*, *question*, and *constraints*. The *context* gives a brief introduction to orient the model with any necessary background information. It can be split into role (e.g. expert annotator) and context (e.g. conspiracy theories). The *question* guides the response, define the coding task. The *constraint* specifies the output format.
- **Give instructions in the correct order**: Recent and repeated text in the prompts has the most effect on LLM generation. It is therefore advisable to start with the *context*, followed by *instructions*, followed by the *constraints*.
- **Enumerate options**: If the answer is categorical, list the options in alphabetical order so that the output is simply the highest-probability token. Each option should be separated by a line-break.
- **Give an "I don't know" option:** Provide an option for the LLM to respond if it is uncertain about the correct answer. This reduces the risk of stochastic answers.
- **Use lists**: If the instruction is complex, make use of explicit lists to help the model pay attention to all elements in the prompt.
- **Use JSON format**: If the answer should contain several pieces of information, request a response in JSON format. The JSON format is easy to parse, and familiar to LLMs.
- **Use an LLM for improving your prompt**: LLMs have been shown to be effective at improving prompts. It can be particularly beneficial to follow an iterative process while utilizing an LLM to provide feedback and produce new versions of a seed prompt (Pryzant et al., 2023).
- **Balance brevity and specificity:** Well-written prompts involve a balance of specificity and brevity. While specificity in a prompt can lead to higher accuracy, performance can fall with longer prompts. Long prompts also make the process more costly, as you will need to feed the prompt for every annotation call.
- **Chain-of-Thought:** For certain tasks, it may be useful to employ more advanced techniques, such as the *Chain-of-Thought* (CoT) technique, to help elicit reasoning in LLMs (Wei, Wang, et al., 2022) and improve instruction-following capabilities (Chung et al., 2022). This involves breaking down the task into several simpler intermediate steps, allowing the LLM to mimic a step-by-step thought process of how humans solve complicated reasoning tasks. It can also be useful to trigger the model to engage in reasoning by using a prefix such as "Let's think step by step."
- **System instructions**: For most LLMs, the prompt instructions are provided as a "system" instruction, with the input as "user" request.
- **Few-shot prompting:** It is often beneficial to also provide examples to guide the desired output, so called *few-shot prompting*, sent as a separate "user" and "assistant" dialogue.

The following provides an example of a well-structured prompt:

> As an expert annotator with a focus on social media content analysis, your role involves scrutinizing Twitter messages related to the US 2020 election. Your expertise is crucial in identifying misinformation that can sway public opinion or distort public discourse.
>
> Does the message contain misinformation regarding the US 2020 election?
>
> Provide your response in JSON format, as follows:
> { "contains_misinformation": "Yes/No/Uncertain", "justification": "Provide a brief justification for your choice." }
>
> Options:
> - Yes
> - No
> - Uncertain
>
> Remember to prioritize accuracy and clarity in your analysis, using the provided context and your expertise to guide your evaluation. If you are uncertain about the classification, choose 'Uncertain' and provide a rationale for this uncertainty.
>
> Twitter message: [MESSAGE]
>
> Answer:

## 6. Specify your LLM parameters

When using an LLM, there are several parameters that can affect the results produced by your prompts. Tweaking these settings are important to improve reliability and desirability of responses, and it may take some experimentation to figure out the appropriate settings for your use cases. The following list shows some common settings you may come across when using LLMs:

- **Max Length** – Sets the maximal number of tokens the model generates. Specifying a max length allows you to control costs, and prevent long or irrelevant responses.

- **Temperature** – The temperature parameter controls how random the model output is, essentially increasing the weights of all other possible tokens. Low temperature leads to more deterministic results, while high temperature leads to more randomness, that is, more diverse or creative outputs. For data annotation, a lower temperature is usually recommended, such as 0.
- **Top-P:** Adjusts the range of considered tokens. A low Top P ensures precise, confident responses, while a higher value promotes diversity by including less likely tokens. For data annotation, a lower Top-P is usually recommended, such as 0.2 to 0.4. If using Top-P, your temperature must be above 0.
- **Top-K:** The top-k parameter limits the model's predictions to the top-k most probable tokens. By setting a value for top-k, you can thereby limit the model to only considering the most likely tokens.

Your parameters *must* always be explicitly specified – even if they are the default parameters – as this is necessary for reproducibility.

### 7. Consider ethical and legal implications

Using LLMs for text analysis opens several ethical considerations compared to traditional text analysis methods, in particular when using platformed LLMs. In regulatory contexts such as the EU, the use of AI furthermore also puts higher legal requirements on data management and ethics.

1. **Transparency and Consent**: Ensure that you have explicit consent from individuals whose personal data you are using that you will employ LLMs for its analysis. Users should be informed about the use of third-party services and the implications for their data. More generally, when using a platformed LLM such as ChatGPT, Claude, or Gemini, your input data is likely to be used as training data.
2. **Data Processing Agreement**: When using third-party services like OpenAI, it may be necessary to have a Data Processing Agreement (DPA) in place. This agreement should outline how the data is processed, the purposes of processing, and the measures taken to protect the data. For instance, if you are using ChatGPT and you are required to be GDPR compliant, you may need to execute a DPA with OpenAI (such an application form is available on the OpenAI website.)

3. **Changing Expectations of Privacy in Public Posts:** The research use of text data that users have published publicly – such as on platforms like X/Twitter or Telegram – is often motivated by users posting such data may have a reduced expectation of privacy. However, the data was likely published without the user considering the substantial capacity of LLMs to extract information, and researchers should thus carefully identify and respect users' expectations of privacy (Zimmer, 2020).
4. **Data Anonymization**: Before sending data to a platformed LLM, ensure that all personal data is adequately anonymized or pseudonymized. This means removing or replacing any information that could directly or indirectly identify an individual. *Never* send proprietary, secret, or confidential data to an API or web interface without careful consideration of the ethical and legal implications.
5. **Data Minimization**: You should only use and send the minimum amount of data necessary. While this is always an important ethical guideline, data minimization is also a legal principle in the European context, as it is part of GDPR.
6. **Data Storage and Transfer**: Be mindful of where the data is stored and processed. The GDPR requires that data transfers outside the EU and the EEA are subject to adequate protections or are made to countries that provide an adequate level of data protection.
7. **Copyright and Terms of Service Violations**: If you are using copyrighted material, such as news articles from a proprietary database, you may need to receive explicit permission or license to analyze the data with an API-based LLM. Without explicit permission or a license from the copyright owner, sending the data to an API can be considered an infringement.

Ethical issues often involve difficult trade-offs. As usual, researchers should handle ethical considerations through an explicit and careful discussion and motivation in their research paper.

## 8. Examine model stochasticity

LLMs behavior in relation to prompts can be brittle and non-intuitive, with even minor details in the prompt – such as capitalization, interpunctuation, or the order of elements or words – significantly impacting accuracy, in some cases even going from state-of-the-art to near random chance (Kaddour et al., 2023; Zhao et al., 2021). Examining whether the model's results are stable can be a useful shortcut to examining whether the model is able to carry out the coding reliably and replicably, without the need for a validation procedure. Does the same prompt return the same result for a given text if run several times? Do small variations in the prompt result in different results? Large variations in output for minor changes in the prompt can indicate issues with the model's stability and reliability for a given task, making its text annotation less trustworthy. If the results are highly sensitive to minor prompt changes, it can also be challenging for other researchers to replicate the study and validate the findings.

To carry out such a prompt stability analysis, create several paraphrases of the prompt, and run the analysis for a subset of the data. You can then estimate the stability by comparing the results, for instance using Krippendorf's Alpha reliability measure (Krippendorff, 2004).

### 9. Consider that your data might be in the training data

When using conventional machine learning models, it is crucial to keep the data you test on separate from the training data to ensure that the model is robust, generalizable, and that it provides a realistic estimate of its performance on unseen data. This may suggest that LLMs cannot be properly validated, as their training data is often so massive that it should be assumed that nearly any publicly available data will be included. However, the general rule does not necessarily apply to LLMs. As the purpose of validating text annotation is to assess the model's capacity for the specific task, it does not matter that the prompt validation data is in the training data, as long as the data on which the model will be run is also in the LLM training data. In fact, it is often desirable that the time-period covered is included in the training data, as it is necessary for the model to draw on contextual knowledge when making inferences about meaning (see Törnberg 2023a). For instance, if the task is to identity the ideology of a poster based on a social media message, it may be necessary to have knowledge of specific policy positions in a given political context.

However, there are situations where this may become problematic. For instance, if *parts* of the text data that you are annotating are in the LLM's training data and other parts are not, the two should preferably be validated separately, as the model's performance may differ. You therefore need to be mindful of the period for which the specific model was trained: if the end date of the LLM training data is within the period of your dataset, you may find that the quality of annotation varies over time – which can cause problems in your downstream analysis.

For the same reason, you should try to avoid using publicly available databases to as validation data, as they may be in the model's training data. For instance, if you are interested in annotating party manifestos, existing manually labelled datasets (such as Manifesto Project Database) are not reliable means of validation: the LLM has likely already seen this database and may simply be reproducing the labels. This implies that the performance may not generalize to tasks for which the answer is not already publicly available. While the risks of such data contamination are often overstated, as the LLMs are trained on massive datasets and is trained as a next-word predictor and may thus be unlikely to have 'memorized' the columns of a CSV file, the burden of evidence is on the validator.

**Conclusion**

This brief paper has collected an emerging set of best practices for text annotation using LLMs, to support both those using the methods as part of their research, and reviewers seeking to evaluate an academic contribution. As the field is undergoing rapid development, it should be noted that the standards and practices should be expected to continue evolving – and this paper is hence far from the final word.

As LLMs fit poorly into our existing epistemic frameworks for text annotation, they have caused a significant academic debate on their role in social scientific research. While many scholars have welcomed the methods – at times with a perhaps overly acritical acclaim – others have rejected them for being unreliable and incompatible with the principles of open science (Kristensen-McLachlan et al., 2023; Ollion et al., 2023, 2024). The suggestion at the core of this paper is that the methods are capable of sophisticated and rigorous interpretation – given appropriate use. LLMs can constitute a powerful contribution to social scientific research, but require a new epistemic apparatus and new standards for evaluating their use.

While critics are largely accurate in describing LLMs as subjective, flawed, black-boxed, potentially biased, and prone to misunderstanding – these descriptions often apply similarly to human coders. To manage these problems, conventional coding is organized in rigorous processes that identify disagreements and validate the reliability. Rather than neither using LLMs uncritically or rejecting them altogether, this implies the possibility to instead structure, direct and formalize their use in ways that harnesses their capacities, while remaining conscious of their inherent weaknesses and risks.

Interpretation is inherently subjective and contested, and the models bring to the surface challenging questions of meaning, nuance, and ambiguity that researchers too often seek to avoid. Qualitative scholars have long known that it is more productive to openly acknowledge and face such issues, rather them concealing them under a veneer of false objectivity. In embracing the subjective and contested nature of interpretation, researchers can engage more deeply with the complexity of their data, acknowledging that different perspectives can lead to differing interpretations. This recognition does not undermine the validity of the research; rather, it enriches the analysis by exposing the multifaceted layers of meaning that exist within the data, and enabling scholars to critically examine their own biases and assumptions. This perspective aligns with the interpretivist paradigm, which posits that reality is socially constructed and that there are multiple, equally valid interpretations of it. Therefore, acknowledging the inherent subjectivity in interpretation is not just a methodological choice but a philosophical stance that values depth, complexity, and the co-construction of meaning between the researcher and the subject.

In employing LLMs for such work, we must be careful to remember that while LLMs can seem in some ways eerily human, they are not human in their capabilities. On some tasks – even those long seen as belonging to the distinctly human realm – they can be superhuman in their interpretive capacities (Törnberg 2023a). On other tasks, they perform worse than a small child. This means that we should not take for granted that their coding matches our intuitive expectations, and that we should always use them carefully and mindfully.

The epistemic challenge that LLMs represent for computational social scientific research is not merely a threat to existing approaches but can productively challenge established conventions. LLMs can empower qualitatively and theoretically minded social scientists to carry out sophisticated computational research, and thus empower a focus on aspects of the social world that have thus been underemphasized in computational research (Törnberg and Uitermark, 2021). Moreover, the models destabilize some of the inequalities of academic research, as a moderately funded early career scholar now can perform analyses that were previously only available to the well-funded lab leader who could afford a team of coders. Such benefits are not to be taken lightly. As Kuhn (1962) famously argued, the most radical scientific advances stem not from accumulated facts and discoveries, but it is the invention of new tools and methodologies that trigger paradigm shifts in scientific work.

**Acknowledgement**

The author wishes to thank Erik Borra of University of Amsterdam for careful comments and feedback on a previous version of this draft, and the all the participants of the workshop *Using LLMs and Text-as-Data in Political Science Research* in Barcelona on January 29, 2024, in particular the organizer Aina Gallega, for their comments, questions and encouragement.

**References**

Alizadeh M, Kubli M, Samei Z, et al. (2023) Open-Source Large Language Models Outperform Crowd Workers and Approach ChatGPT in Text-Annotation Tasks. arXiv:2307.02179. arXiv. Available at: http://arxiv.org/abs/2307.02179 (accessed 23 January 2024).

Bommasani R, Liang P and Lee T (2023) Holistic Evaluation of Language Models. *Annals of the New York Academy of Sciences* 1525(1): 140–146.

Chang Yupeng, Wang X, Wang J, et al. (2024) A Survey on Evaluation of Large Language Models. *ACM Transactions on Intelligent Systems and Technology*: 3641289.

Chen L, Zaharia M and Zou J (2023) How is ChatGPT's behavior changing over time? arXiv:2307.09009. arXiv. Available at: http://arxiv.org/abs/2307.09009 (accessed 4 February 2024).

Chia YK, Hong P, Bing L, et al. (2023) INSTRUCTEVAL: Towards Holistic Evaluation of Instruction-Tuned Large Language Models. arXiv:2306.04757. arXiv. Available at: http://arxiv.org/abs/2306.04757 (accessed 4 February 2024).

Chun Tie Y, Birks M and Francis K (2019) Grounded theory research: A design framework for novice researchers. *SAGE Open Medicine* 7: 2050312118822927.

Chung HW, Hou L, Longpre S, et al. (2022) Scaling Instruction-Finetuned Language Models. arXiv:2210.11416. arXiv. Available at: http://arxiv.org/abs/2210.11416 (accessed 5 February 2024).

Fernandes P, Madaan A, Liu E, et al. (2023) Bridging the Gap: A Survey on Integrating (Human) Feedback for Natural Language Generation. arXiv:2305.00955. arXiv. Available at: http://arxiv.org/abs/2305.00955 (accessed 5 February 2024).

Gilardi F, Alizadeh M and Kubli M (2023) ChatGPT outperforms crowd workers for text-annotation tasks. *Proceedings of the National Academy of Sciences* 120(30). Proceedings of the National Academy of Sciences: e2305016120.

Glaser BG and Strauss AL (2009) *The Discovery of Grounded Theory: Strategies for Qualitative Research*. New Brunswick: Aldine.

HuggingFace (2024) Open LLM Leaderboard. Available at: https://huggingface.co/spaces/HuggingFaceH4/open_llm_leaderboard (accessed 4 February 2024).

Jiang Z, Xu FF, Araki J, et al. (2020) How can we know what language models know? *Transactions of the Association for Computational Linguistics* 8. MIT Press One Rogers Street, Cambridge, MA 02142-1209, USA journals-info …: 423–438.

Kaddour J, Harris J, Mozes M, et al. (2023) Challenges and Applications of Large Language Models. arXiv:2307.10169. arXiv. Available at: http://arxiv.org/abs/2307.10169 (accessed 5 February 2024).

Karjus A (2023) Machine-assisted mixed methods: augmenting humanities and social sciences with artificial intelligence. arXiv:2309.14379. arXiv. Available at: http://arxiv.org/abs/2309.14379 (accessed 5 February 2024).

Krippendorff K (2004) Reliability in content analysis: Some common misconceptions and recommendations. *Human communication research* 30(3). Wiley Online Library: 411–433.

Kristensen-McLachlan RD, Canavan M, Kardos M, et al. (2023) Chatbots Are Not Reliable Text Annotators. arXiv:2311.05769. arXiv. Available at: http://arxiv.org/abs/2311.05769 (accessed 23 January 2024).

Kuhn TS (1962) *The structure of scientific revolutions.* Chicago: University of Chicago Press

Liesenfeld A, Lopez A and Dingemanse M (2023) Opening up ChatGPT: Tracking openness, transparency, and accountability in instruction-tuned text generators. In: *Proceedings of the 5th International Conference on Conversational User Interfaces*, Eindhoven Netherlands, 19 July 2023, pp. 1–6. ACM. Available at: https://dl.acm.org/doi/10.1145/3571884.3604316 (accessed 4 February 2024).

Ollion E, Shen R, Macanovic A, et al. (2023) Chatgpt for Text Annotation? Mind the Hype! SocArXiv. https://osf.io/preprints/socarxiv/x58kn

Ollion É, Shen R, Macanovic A, et al. (2024) The dangers of using proprietary LLMs for research. *Nature Machine Intelligence*. Nature Publishing Group UK London: 1–2.

OpenAI (2024) Prompt engineering. Available at: https://platform.openai.com/docs/guides/prompt-engineering/strategy-write-clear-instructions (accessed 5 February 2024).

Pangakis N, Wolken S and Fasching N (2023) Automated Annotation with Generative AI Requires Validation. arXiv:2306.00176. arXiv. Available at: http://arxiv.org/abs/2306.00176 (accessed 23 January 2024).

Pryzant R, Iter D, Li J, et al. (2023) Automatic Prompt Optimization with 'Gradient Descent' and Beam Search. arXiv:2305.03495. arXiv. Available at: http://arxiv.org/abs/2305.03495 (accessed 5 February 2024).

Rathje S, Mirea D-M, Sucholutsky I, et al. (2023) GPT is an effective tool for multilingual psychological text analysis. PsyArXiv. Epub ahead of print 2023.

Sanh V, Webson A, Raffel C, et al. (2022) Multitask prompted training enables zero-shot task generalization. In: *International Conference on Learning Representations*, 2022. Available at: https://iris.uniroma1.it/handle/11573/1672126 (accessed 5 February 2024).

Saravia E (2022) Prompt Engineering Guide. Available at: https://github.com/dair-ai/Prompt-Engineering-Guide (accessed 4 February 2024).

Spirling A (2023) World view. *Nature* 616: 413.

Törnberg P (2023) Chatgpt-4 outperforms experts and crowd workers in annotating political twitter messages with zero-shot learning. *arXiv preprint arXiv:2304.06588*. Epub ahead of print 2023.

Törnberg P (2024) How to Use Large-Language Models for Text Analysis. SAGE Publications Ltd. Epub ahead of print 2024.

Törnberg P and Uitermark J (2021) For a heterodox computational social science. *Big Data & Society* 8(2). SAGE Publications Sage UK: London, England: 20539517211047725.

Weber M and Reichardt M (2023) Evaluation is all you need. Prompting Generative Large Language Models for Annotation Tasks in the Social Sciences. A Primer using Open Models. arXiv:2401.00284. arXiv. Available at: http://arxiv.org/abs/2401.00284 (accessed 23 January 2024).

Wei J, Wang X, Schuurmans D, et al. (2022) Chain-of-thought prompting elicits reasoning in large language models. *Advances in Neural Information Processing Systems* 35: 24824–24837.

Wei J, Bosma M, Zhao VY, et al. (2022) Finetuned Language Models Are Zero-Shot Learners. arXiv:2109.01652. arXiv. Available at: http://arxiv.org/abs/2109.01652 (accessed 4 February 2024).

White J, Hays S, Fu Q, et al. (2023) ChatGPT Prompt Patterns for Improving Code Quality, Refactoring, Requirements Elicitation, and Software Design. arXiv:2303.07839. arXiv. Available at: http://arxiv.org/abs/2303.07839 (accessed 23 January 2024).

Yu H, Yang Z, Pelrine K, et al. (2023) Open, Closed, or Small Language Models for Text Classification? arXiv:2308.10092. arXiv. Available at: http://arxiv.org/abs/2308.10092 (accessed 23 January 2024).
20

Zhao Z, Wallace E, Feng S, et al. (2021) Calibrate before use: Improving few-shot performance of language models. In: *International Conference on Machine Learning*, 2021, pp. 12697–12706. PMLR. Available at: http://proceedings.mlr.press/v139/zhao21c.html (accessed 5 February 2024).

Ziegler DM, Stiennon N, Wu J, et al. (2019) Fine-tuning language models from human preferences. *arXiv preprint arXiv:1909.08593*. Epub ahead of print 2019.

Zimmer M (2020) "But the data is already public": on the ethics of research in Facebook. In: *The Ethics of Information Technologies*. Routledge, pp. 229–241. Available at: https://www.taylorfrancis.com/chapters/edit/10.4324/9781003075011-17/data-already-public-ethics-research-facebook-michael-zimmer (accessed 5 February 2024).